\documentclass{article}

\usepackage[numbers]{natbib}
\usepackage[preprint]{neurips_2023}

\usepackage{amsmath,amsfonts,bm}

\def\eqref#1{equation~(\ref{#1})}
\def\Eqref#1{Equation~(\ref{#1})}

\def\1{\bm{1}}

\DeclareMathAlphabet{\mathsfit}{\encodingdefault}{\sfdefault}{m}{sl}
\SetMathAlphabet{\mathsfit}{bold}{\encodingdefault}{\sfdefault}{bx}{n}

\newcommand{\R}{\mathbb{R}}

\usepackage[dvipsnames]{xcolor}         
\definecolor{linkColor}{rgb}{0.18,0.39,0.62}
\usepackage[utf8]{inputenc} 
\usepackage[T1]{fontenc}    
\usepackage[colorlinks=true,linkcolor=linkColor,citecolor=linkColor,filecolor=linkColor,urlcolor=linkColor]{hyperref}       
\usepackage{multirow}
\usepackage{booktabs}
\usepackage{graphicx}
\usepackage{wrapfig}
\usepackage{bm}
\usepackage{algorithm}
\usepackage{algpseudocode}
\usepackage{enumitem}
\usepackage{tcolorbox}
\usepackage{makecell}
\usepackage{diagbox}

\usepackage{amsmath}
\usepackage{amssymb}
\usepackage{mathtools}
\usepackage{amsthm}

\def\expct{\mathop{\mathbb{E}}}

\def\pst{q_{\theta}(y_t | \by_{<t}, \bx)}
\def\pstp{q_{\theta}(y_{t'} | \by_{<t'}, \bx)}

\def\ptt{p(y_t | \by_{<t}, \bx)}
\def\pttp{p(y_{t'} | \by_{<t'}, \bx)}
\def\pmt{\widetilde{p}(y_t | \by_{<t}, \bx)}

\def\forkl{\operatorname{KL}[p||q_\theta]}
\def\revkl{\operatorname{KL}[q_\theta||p]}
\def\ps{q_\theta}
\def\pt{p}
\def\bx{\bm{x}}
\def\by{\bm{y}}
\def\R{\log \frac{\pt (\by | \bx)}{\ps (\by | \bx)}}
\def\dy{\mathrm{d} \by}
\def\dx{\mathrm{d} \bx}

\title{MiniLLM: On-Policy Distillation of Large Language Models}

\author{%
  Yuxian Gu$^{1,2}$\thanks{Contribution during an internship at Microsoft Research.},~~~\ Li Dong$^2$,~~~\ Furu Wei$^2$,~~~\ Minlie Huang$^1$\thanks{Corresponding author.} \\
  $^1$The CoAI Group, Tsinghua University\\
  $^2$Microsoft Research \\
  \small \texttt{guyx21@mails.tsinghua.edu.cn} \ \ \ \texttt{\{lidong1,fuwei\}@microsoft.com}\\
  \small \texttt{aihuang@tsinghua.edu.cn}\\
}

\begin{document}

\maketitle

\begin{abstract}
Knowledge Distillation (KD) is a promising technique for reducing the high computational demand of large language models (LLMs). However, previous KD methods are primarily applied to white-box classification models or training small models to imitate black-box model APIs like ChatGPT. How to effectively distill the knowledge of white-box LLMs into small models is still under-explored, which becomes more important with the prosperity of open-source LLMs. In this work, we propose a KD approach that distills LLMs into smaller language models. We first replace the \textit{forward} Kullback-Leibler divergence (KLD) objective in the standard KD approaches with \textit{reverse} KLD, which is more suitable for KD on generative language models, to prevent the student model from overestimating the low-probability regions of the teacher distribution. Then, we derive an effective \textit{on-policy} optimization approach to learn this objective. The student models are named \textbf{\textsc{MiniLLM}}. Extensive experiments in the instruction-following setting show that \textsc{MiniLLM} generates more precise responses with higher overall quality, lower exposure bias, better calibration, and higher long-text generation performance than the baselines. Our method is scalable for different model families
with 120M to 13B parameters. Our code, data, and model checkpoints can be found in \url{https://github.com/microsoft/LMOps/tree/main/minillm}.
\end{abstract}

\begin{figure}[H]
\centering
\includegraphics[width=\textwidth]{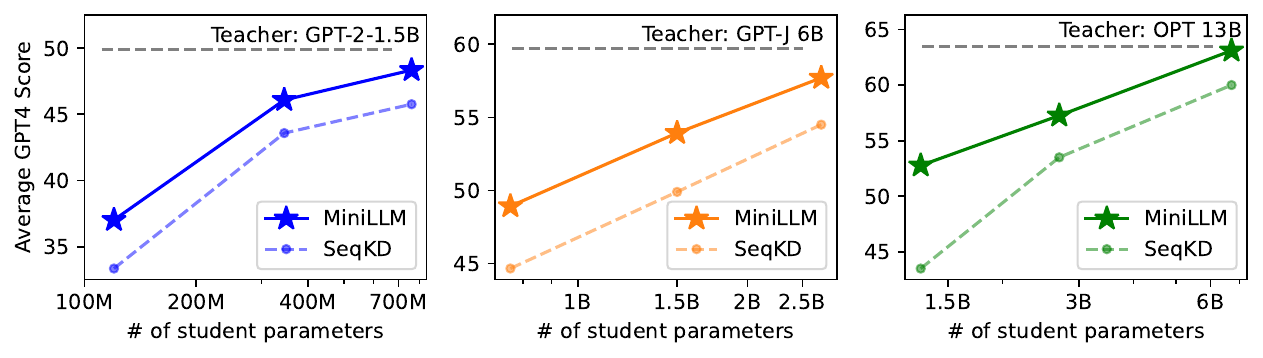}
\caption{The comparison of \textsc{MiniLLM} with the sequence-level KD (SeqKD;~\citealp{skd,alpaca,vicuna,ITGPT4,false_imitate,lima}) in terms of the average GPT-4 feedback score on our evaluation sets. \textbf{Left}: GPT-2-1.5B as the teacher model and GPT-2 125M, 340M, 760M as the student models. \textbf{Middle}: GPT-J 6B as the teacher model and GPT-2 760M, 1.5B, GPT-\textit{Neo} 2.7B as the student models. \textbf{Right}: OPT 13B as the teacher and OPT 1.3B, 2.7B, 6.7B as the student models.}
\label{fig:full_res}
\end{figure}

\newpage
\section{Introduction}

With the rapid development of large language models (LLMs;~\citealp{gpt3,plmsurvey,foundation_model,palm,gpt4}),
a common technique to reduce their high computational resource demand is knowledge distillation (KD;~\citealp{kd}), where we train a small student model with supervision from a large teacher model. Two categories of KD are commonly applied: \textit{black-box} KD, where only the teacher-generated texts are accessible, and \textit{white-box} KD, where the teacher model's output distribution or intermediate hidden states are also available~\citep{kdsurvey}.
Recently, \textit{black-box} KD has shown promising results in fine-tuning small models on the prompt-response pairs generated by LLM APIs~\citep{alpaca, vicuna,lamini, ITGPT4}. With the emergence of more open-source LLMs~\citep{opt, llama}, \textit{white-box} KD becomes more valuable for both research communities and industry sectors because student models receive better signals from the output distribution and hidden states of teacher models, thereby potentially resulting in higher performance. However, \textit{white-box} KD approaches are mostly studied for small ($<$ 1B parameters) language understanding models~\citep{distilbert,minilm}, while \textit{white-box} KD for LLMs is yet to be explored.

In this work, we investigate \textit{white-box} KD of LLMs where the output distribution of the teacher model is available. We argue that the standard KD objectives~\citep{skd,lightpaff,vicuna,alpaca} are sub-optimal for LLMs that perform tasks in a generative manner. Given the teacher distribution $\pt(\by|\bx)$ and the student distribution $\ps(\by|\bx)$ parameterized by $\theta$, standard KD objectives (including several variants for sequence-level models) essentially minimize the approximated \textit{forward} Kullback-Leibler divergence (KLD) between the teacher and the student distribution, termed as $\forkl$, which forces $\ps$ to cover all modes of $\pt$. For text classification tasks, $\forkl$ works well because the output space usually consists of a finite number of classes such that both $\pt(\by|\bx)$ and $ \ps(\by|\bx)$ have few modes. However, for open-ended text generation tasks, which is usually the case of LLM applications, the output spaces are much more complex and $\pt(\by|\bx)$ can contain many more modes than what $\ps(\by|\bx)$ can express due to the limited model capacity. Minimizing \textit{forward} KLD causes $\ps$ to assign unreasonably high probabilities to the void regions of $\pt$~\citep{malinin2019reverse} and produces very unlikely samples under $\pt$ during free-run generation~\citep{huszar2015not}.

\begin{wrapfigure}{r}{5.5cm}
    \centering
    \vspace{-0.45cm}
    \includegraphics[width=0.35\textwidth]{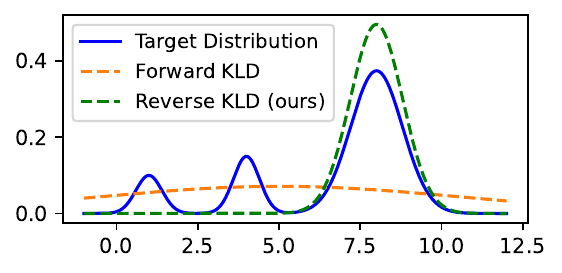}
    \vspace{-0.25cm}
    \caption{The toy experiment. We fit a Gaussian mixture distribution with a single Gaussian distribution using \textit{forward} KLD and \textit{reverse} KLD.}
    \vspace{-0.1cm}
    \label{fig:toy}
\end{wrapfigure}
To alleviate this problem, we propose to minimize \textit{reverse} KLD, $\revkl$, widely used in computer vision~\citep{selfkd} and reinforcement learning~\citep{kd_policy_kd}. Compared to $\forkl$, minimizing $\revkl$ causes $\ps$ to seek the major modes of $\pt$, and assign low probabilities to $\pt$'s void regions~\citep{divergence_measure}, as illustrated in Figure \ref{fig:toy} and discussed in Section \ref{sec:minillm}. In text generation, this means that the student avoids learning too many long-tail variants~\citep{top_p} in the teacher's distribution to focuses on the generation correctness, which is critical in practical scenarios that require truthfulness and reliability~\citep{halu_survey}. To optimize $\min_\theta \revkl$, as shown in Section~\ref{sec:rl_optim}, we derive its gradient with Policy Gradient~\citep{policy_gradient} and adopt an \textit{on-policy} training approach.
To further stabilize and accelerate training, we propose (1) single-step decomposition to reduce variance, (2) teacher-mixed sampling to alleviate reward hacking,
and (3) length normalization to eliminate the length bias. Finally, we introduce the overall KD algorithm in Section~\ref{sec:alg}.  Our student models are named \textsc{\textbf{MiniLLM}}, indicating our method is suitable for compressing large (generative) language models.

We apply our method to various generative language models~\citep{gpt2,opt,llama} with sizes ranging from 120M to 13B in the instruction-following setting~\citep{t0,flan} that covers a large range of NLP tasks. We use 5 datasets with Rouge-L~\citep{rouge}, the GPT-4 feedback, and human judgment for evaluation. Experiments show that \textsc{MiniLLM} consistently outperforms standard KD baselines on all the datasets and scales up well from 120M to 13B models (see Figure \ref{fig:full_res}). More analysis shows that \textsc{MiniLLM} yields lower exposure bias, better calibration, and higher long response generation performance, with neglectable loss of diversity.

\section{Method}
\label{sec:method}
\vspace{-0.1cm}
We consider conditional text generation where the model produces a response $\by=\{y_t\}_{t=1}^T$ conditioning on a prompt $\bx$ sampled from the distribution $p_{\bm{x}}$, which is typically how LLMs perform tasks. We formulate KD as an optimization problem to minimize the difference between a fixed teacher model distribution $p(\by|\bx)$ and a student model distribution $q_\theta(\by|\bx)$ parameterized by $\theta$. The standard KD methods approximately\footnote{We say ``approximately'' because for word-level KD, $\by$ is sampled from the real distribution, not the teacher distribution. For a strong enough teacher model, we can consider the two distributions approximately the same.} minimize the \textit{forward} KLD: $\forkl = \expct_{\bx \sim p_{\bx}, \by \sim p'} \log \frac{p(\by|\bx)}{q_\theta (\by|\bx)}$, where $p'$ can be real data distribution (word-level KD) or teacher distribution $p$ (sequence-level KD). Though widely used, $\forkl$ tends to overestimate the void regions of $p$ in text generation tasks when $q_\theta$ is insufficiently expressive~\citep{tailor}. KD for LLMs fits the case because LLMs perform tasks in a generative manner, such that the low-capacity student models cannot perfectly imitate the complex text generation distribution of the teacher models or humans.

\subsection{\textsc{MiniLLM}: Knowledge Distillation with \textit{Reverse} KLD}
\label{sec:minillm}

\begin{figure}[t]
\centering
\includegraphics[width=0.89\linewidth]{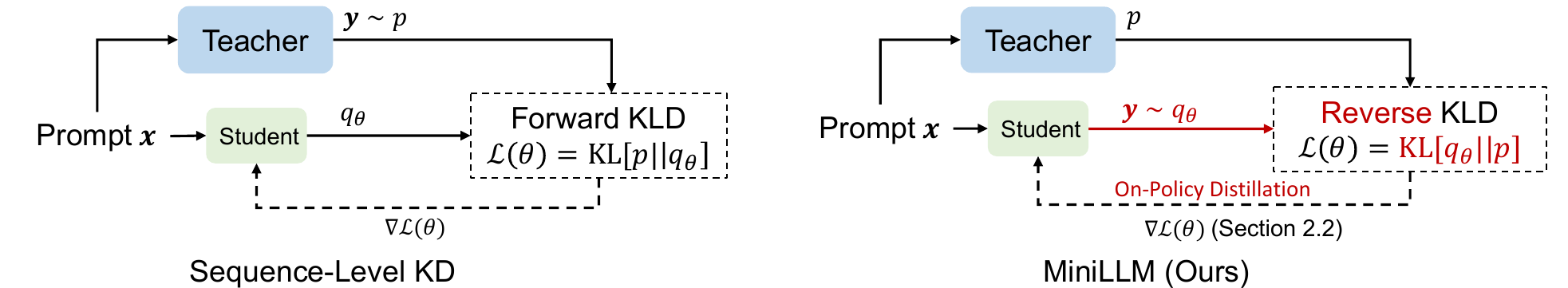}
\vspace{-0.1cm}
\caption{Comparison between sequence-level KD (left) and \textsc{MiniLLM} (right). Sequence-level KD forces the student to memorize all samples generated by the teacher model, while \textsc{MiniLLM} improves its generated texts with the teacher model's feedback.}
\vspace{-0.1cm}
\label{fig:method}
\end{figure}

We consider minimizing the \textit{reverse} KLD between the student and teacher model distributions as the learning objective for \textsc{MiniLLM}:
\begin{equation}
\small
\begin{aligned}
\label{eq:obj}
\theta = \arg \min_\theta \mathcal{L}(\theta) &= \arg \min_\theta \revkl \\
&= \arg \min_\theta \left[-\expct_{\bx \sim p_{\bx},\by \sim q_\theta} \log \frac{p (\by|\bx)}{q_\theta(\by|\bx)}\right].
\end{aligned}
\end{equation}
Minimizing \textit{reverse} KLD has been shown to cause the mode-seeking behavior in generative modeling~\citep{huszar2015not,f-gan,chen2018symmetric,selfkd}, where $q_\theta$ assigns high probabilities to $p$'s large modes and ignore the small ones (illustrated in a toy experiment in Figure \ref{fig:toy}). In this work, we first study this property for KD of LLMs in text generation.
Minimizing \textit{forward} KLD causes $q_\theta$ to place large probability masses on the zero-probability regions of $p$, corresponding to the generation of low-quality text in practice, while \textit{reverse} KLD focuses on $p$'s major modes, which is crucial to ensure the correctness and faithfulness of text generation. As illustrated in Figure \ref{fig:method}, unlike sequence-level KD that minimizes \textit{forward} KLD~\cite{skd,alpaca}, \textsc{MiniLLM} that minimizes \textit{reverse} KLD does not force $\ps$ to fit all $\by$ sampled from the teacher distribution $p$. Instead, it encourages the student to generate samples preferred by the teacher within its own capacities, which is more possible to achieve. Interestingly, we also find another perspective of understanding \textsc{MiniLLM} motivated by Inverse Reinforcement Learning~\citep{maxent_irl}. We present the related derivation in Appendix \ref{app:persp_rl}.

\subsection{On-Policy Distillation}

\label{sec:rl_optim}

\vspace{-2pt}

\paragraph{Gradient Derivation} 
We notice that the gradient of the objective function $\mathcal{L}(\theta)$ in \Eqref{eq:obj} can be derived using the Policy Gradient Theorem~\citep{reinforce,max_ent_rl} for \textit{on-policy} optimization:
\begin{equation}
    \small
    \label{eq:grad}
    \nabla \mathcal{L}(\theta) = -\expct_{\bx \sim p_{\bx}, \by \sim q_{\theta}(\cdot | \bx)} \sum_{t=1}^T (R_t-1) \nabla \log \pst,
\end{equation}
where $T=|\by|$ and $R_t =  \sum_{t'=t}^T \log \frac{\pttp}{\pstp}$ is the accumulation of $r_{t'}=\log \frac{\pttp}{\pstp}$ that measures the quality of each step's generation. Intuitively, the generated texts are supposed to have high probabilities under the teacher distribution by increasing $\pttp$, but simultaneously stay diverse by lowering $\pstp$. The expectation in Eq. \ref{eq:grad} is computed by Monte-Carlo sampling. Full derivation can be found in Appendix \ref{app:grad}. However, policy gradient suffers from high variance and reward hacking~\citep{reward_hacking}, despite some subsequent solutions~\citep{ppo}. Besides, we notice that $R_t$ favors short sentences, which causes the student model to output empty responses. Therefore, we propose three strategies to mitigate these problems.

\vspace{-2pt}

\paragraph{Single-Step Decomposition}
\cite{kd_policy_kd} has found that the single-step generation quality $r_t$ is critical to the training variance because the error in the front tokens accumulates along the whole sentence. To pay more attention to $r_t$, we re-write $\nabla \mathcal{L}(\theta)$ to decompose $r_t$ from $R_t$ and directly compute the gradient of $\expct_{y_t \sim q_\theta (t)}[r_t]$ (see Appendix \ref{app:exp_reg} for the full derivation):
\begin{equation}
\small
\begin{aligned}
    \label{eq:exp_reg}
    \nabla \mathcal{L}(\theta) &= \expct_{\substack{\bx \sim p_{\bx} \\ \by \sim q_{\theta}(\cdot | \bx)}} \left[ - \sum_{t=1}^T\nabla \expct_{y_t \sim q_\theta (t)}[r_t]\right] + \expct_{\substack{\bx \sim p_{\bx} \\ \by \sim q_{\theta}(\cdot | \bx)}} \left[ -\sum_{t=1}^T R_{t+1} \nabla \log \pst \right]\\
    &= (\nabla \mathcal{L})_{\text{Single}} + (\nabla \mathcal{L})_{\text{Long}},
\end{aligned}    
\end{equation}

where $q_\theta(t)=q_\theta(\cdot | \by_{<t}, \bx)$. 
Note that $\expct_{y_t \sim q_\theta(t)}[r_t]$ can be computed directly by summing over the vocabulary instead of using Monte-Carlo sampling and is derivable with respect to $\theta$. This decomposition gives a more precise and efficient estimation of the single-step generation quality, which reduces the variance during training and accelerates convergence.

\vspace{-2pt}

\paragraph{Teacher-Mixed Sampling}
We observe reward hacking~\citep{reward_hacking} when training with Eq. \ref{eq:grad} because $q_\theta$ sometimes produces degenerated sentences $\by$ that receive high scores from the teacher (e.g., repeated phrases) during sampling, especially for small student models. To create a better sampling distribution, we mix the teacher and the student distribution at each time step:
\begin{equation}
\small
\label{eq:teacher_mix}
 \pmt  = \alpha \cdot \ptt + (1-\alpha) \cdot \pst,
\end{equation}
where $\alpha$ controls the strength of the teacher mix-in. Sampling from $\widetilde{p}$ suppresses low-quality generation with the teacher's help and alleviates reward hacking. We re-write $(\nabla \mathcal{L})_{\text{Single}}$ and $(\nabla \mathcal{L})_{\text{Long}}$ with importance sampling to get to an unbiased estimator of the gradient~\citep{off_policy}:
\begin{equation}
    \small
    \label{eq:off-policy}
    \begin{aligned}
        (\nabla \mathcal{L})_{\text{Single}} &= -\expct_{\bx \sim p_{\bx}, \by \sim \widetilde{p}(\cdot | \bx)} \left[ \sum_{t=1}^T w_t \nabla \expct_{y_t \sim q_\theta (t)}[r_t]\right], \\
        (\nabla \mathcal{L})_{\text{Long}} &= -\expct_{\bx \sim p_{\bx}, \by \sim \widetilde{p}(\cdot | \bx)} \left[\sum_{t=1}^T w_tR_{t+1} \nabla \log \pst\right],
    \end{aligned}
\end{equation}
where $w_t=\prod_{t'=1}^{t} \frac{q_\theta(y_{t'}|\by_{<t'}, \bx)}{\widetilde{p}(y_{t'} | \by_{<t'}, \bx)}$ is the importance weight. However, $w_t$ brings high variance in practice because it requires multiplying per-token importance weight over multiple time steps, and thus the variance of each step accumulates. Therefore, we approximately set $w_t \approx \frac{\pst}{\widetilde{p}(y_{t} | \by_{<t}, \bx)}$ to reduce the variance of the estimator in Eq.~\ref{eq:off-policy}~\citep{deep_rl_chat,offline_rl}.

\vspace{-2pt}

\paragraph{Length Normalization}
We found that long sequences tend to have small $R_{t+1}$, which encourages the model to produce short responses. Therefore, we add length normalization to $R_{t+1}$ in Eq.~\ref{eq:exp_reg}: 
\begin{equation}
\small
\label{eq:q_norm}
R^{\text{Norm}}_{t+1} = \frac{1}{T-t-1}\sum_{t'=t+1}^T \log \frac{\pttp}{\pstp}. 
\end{equation}

\paragraph{In Summary}
Combining the strategies listed above, we have the final optimization gradient:
\begin{equation}
\delimiterfactor=1500
\delimitershortfall=20pt
    \small
    \nabla \mathcal{L}(\theta) = -\expct_{\substack{\bx \sim p_{\bx} \\ \by \sim \widetilde{p}(\cdot | \bx)}} \Big[ \sum_{t=1}^T w_t\Big[ \underbrace{\nabla \sum_{y' \in V} q_\theta (y'|\by_{<t}, \bx) \log \frac{p (y'|\by_{<t}, \bx)}{q_\theta (y'|\by_{<t}, \bx)}}_{(\nabla \mathcal{L})_{\text{Single}} \text{ part}} + \underbrace{ \vphantom{ \sum_{y' \in V} } R_{t+1}^{\text{Norm}} \frac{\nabla \pst}{\pst}}_{ (\nabla \mathcal{L})_{\text{Long}}^{\text{Norm}} \text{ part}}\Big]\Big],
\end{equation}
where $V$ is the vocabulary size of the language model and $(\nabla \mathcal{L})_{\text{Long}}^{\text{Norm}}$ is $(\nabla \mathcal{L})_{\text{Long}}$ with $R^{\text{Norm}}_{t+1}$.


\subsection{Training Algorithm}
\label{sec:alg}
We start from a student model pre-trained on a large long-document corpus $\mathcal{D}_{\text{PT}}$. Algorithm \ref{alg:minillm} trains \textsc{MiniLLM} by adapting the student model to a text generation task with dataset $\mathcal{D}$ and supervision from the teacher model, such as an LLM fine-tuned on $\mathcal{D}$~\citep{alpaca,vicuna} or that with good task-generalization~\citep{flan-t5,gpt4}. In the training algorithm, we first fine-tune the student model on $\mathcal{D}$ and pick the checkpoint with the lowest loss as an initialization for the following training. Then, we compute the gradients $(\nabla \mathcal{L})_{\text{Single}}$ and $(\nabla \mathcal{L})_{\text{Long}}^{\text{Norm}}$ based on Eq. \ref{eq:off-policy} and Eq. \ref{eq:q_norm}, with a clipping strategy~\citep{ppo} added to further improve stability.
Same as \cite{instruct-gpt}, we include a language modeling loss $\mathcal{L}_{\text{PT}} = -\expct_{\bm{d} \sim \mathcal{D}_\text{PT}} \log q_\theta (\bm{d})$ to preserve the model performance on canonical NLP benchmarks. The student model is finally updated using a combination of gradients $(\nabla \mathcal{L})_{\text{Single}} + (\nabla \mathcal{L})_{\text{Long}}^{\text{Norm}} + \nabla \mathcal{L}_\text{PT}$. The whole on-policy training pipeline is similar to Reinforcement Learning from Human Feedback (RLHF;~\citealp{instruct-gpt}). 

\begin{algorithm}
    \small
    \label{alg:minillm}
    \caption{\textsc{MiniLLM}: Knowledge Distillation of LLMs}
    \begin{algorithmic}
        \Require Conditional generation dataset $\mathcal{D}$ consisting of prompts and ground-truth responses
        \State \quad \quad \ Pre-training corpus $\mathcal{D}_{\text{PT}}$ consisting of long-document plain texts
        \State \quad \quad \ A teacher model with output distribution $p$
        \State \quad \quad \ An initial student model pre-trained on $\mathcal{D}_{\text{PT}}$, with the output distribution $q_{\theta_0}$
        \State \quad \quad \ Learning rate $\eta$; \quad Batch size $M$; \quad Clipping Threshold $\epsilon$
        \Ensure A student model with the output distribution $q_\theta$
        \State Fine-tune the student model from $\theta_0$ on $\mathcal{D}$ supervised by the ground truth responses and choose $\theta$ with the lowest validation loss.
        \Repeat
        \State Sample a mini-batch of prompts from $\mathcal{D}$ and collect responses from $\widetilde{p}$ to get $\mathcal{S}=\{(\bx^m, \by^m)\}_{m=1}^M$ 
        \State Sample a mini-batch $\mathcal{D'}_{\text{PT}} = \{\bm{d}^m\}_{m=1}^M$ from $\mathcal{D}_{\text{PT}}$
        \State Compute 
        $
            (\nabla \mathcal{L})_{\text{Single}} = -\frac{1}{M}\sum_{\bx, \by\in \mathcal{S}} \sum_{t=1}^T w_t \nabla \sum_{y_t \in V} \pst \log \frac{\ptt}{\pst} 
        $
        \Comment{Eq. \ref{eq:off-policy}}
        \State Compute 
        $
            (\nabla \mathcal{L})_{\text{Long}}^{\text{Norm}}=-\frac{1}{|M|}\sum_{\bx, \by\in \mathcal{S}} \sum_{t=1}^T R^{\text{Norm}}_{t+1}\nabla \min [\rho_t(\theta), \operatorname{clip}(\rho_t(\theta), 1-\epsilon, 1+\epsilon)],
        $
        \State where $\rho_t(\theta) = \frac{\pst}{\pmt}$ \Comment{Eq. \ref{eq:off-policy}, Eq. \ref{eq:q_norm}}
        \State{Compute the gradient of the language modeling loss:}
        $
            \nabla \mathcal{L}_\text{PT} = -\frac{1}{M} \sum_{\bm{d} \in D'_\text{PT}} \nabla \log q_{\theta}(\bm{d})
        $
        \State{Update model parameters}: $\theta \leftarrow \theta - \eta\left[(\nabla \mathcal{L})_{\text{Single}} + (\nabla \mathcal{L})_{\text{Long}}^{\text{Norm}} + \nabla \mathcal{L}_\text{PT}\right]$
        \Until{converge} and \Return $q_\theta$
    \end{algorithmic}
\end{algorithm}

\section{Experiments}

\subsection{Experimental Setup}
We take instruction-following~\citep{instruct-gpt} as the conditional text generation task, where models are trained to generate responses according to the instructions. We fine-tune a large model on the dataset $\mathcal{D}$ consisting of instruction-response pairs as the teacher model. Then, we compare different KD methods on $\mathcal{D}$ by evaluating the student model's instruction-following performance.

\paragraph{Base Models} Our student models come from three model families with various sizes: GPT-2~\citep{gpt2} (120M, 340M, 760M), OPT~\citep{opt} (1.3B, 2.7B, 6.7B), and LLaMA~\citep{llama} (7B). For teacher models of each model family, we use GPT-2-1.5B, OPT-13B, and LLaMA-13B respectively. These models are fine-tuned on $\mathcal{D}$ in advance. We also present the results using GPT-J~\citep{gpt-j} as the teacher model in Appendix \ref{app:res_gptj}.

\paragraph{Training} We construct the training data from \texttt{databricks-dolly-15K}\footnote{\url{https://github.com/databrickslabs/dolly/tree/master}} consisting of 15K human-written instruction-response pairs. We filter out samples that exceed the context length of the models. Then, we randomly split 1K and 0.5K samples for validation and testing, respectively, leaving about 12.5K examples for training. For $\mathcal{D}_{\text{PT}}$, we use OpenWebText~\citep{openwebtext} for the GPT-2 family and the RoBERTa training corpus~\citep{roberta} for other models. We set the teacher-mix-in strength $\alpha=0.2$ throughout the experiments in Eq. \ref{eq:teacher_mix}. We use Rouge-L~\citep{rouge} scores on the validation set to search for hyper-parameters because it aligns better with human preference than validation losses~\citep{super-natural-instructions}. More details are shown in Appendix \ref{app:training_detail}.

\paragraph{Evaluation} We evaluate the trained models on five instruction-following datasets: 
\begin{itemize}[leftmargin=12pt,noitemsep,topsep=-3pt]
    \item \textbf{DollyEval}: the 500-sample test set we split from the \texttt{databricks-dolly-15k} dataset. 
    \item \textbf{SelfInst}~\citep{self_inst}: A user-oriented instruction-following set with 252 samples.
    \item \textbf{VicunaEval}~\citep{vicuna}: The 80 challenging questions used in the Vicuna evaluation. 
    \item \textbf{S-NI}: The test set of \textsc{Super-NaturalInstructions}~\citep{super-natural-instructions} consisting of 9K samples ranging from 119 tasks. Following~\cite{ITGPT4}, we split the set into 3 subsets whose ground truth response lengths lie in $[0,5]$, $[6,10]$, and $[11,+\infty]$. We use the $[11, +\infty]$ subset in Section~\ref{sec:auto} and conduct an analysis on all subsets in Section~\ref{sec:ana}.
    \item \textbf{UnNI}: We randomly sample 10K samples from the core set of \textsc{UnnaturalInstructions}~\citep{uinst} for evaluation. Similar to \textbf{S-NI}, we first conduct the evaluations on the $[11, +\infty]$ subset, followed by an analysis of the performance on all subsets in Appendix \ref{app:length_uni}.
\end{itemize}
We adopt three metrics to evaluate the model-generated responses: 
\begin{itemize}[leftmargin=12pt,topsep=-1pt]
    \item \textbf{R-L}: The Rouge-L~\citep{rouge} score to measure the precision of the model generation. \cite{super-natural-instructions} has shown that Rouge-L is suitable for large-scale instruction-following evaluation.
    \item \textbf{GPT4}: The GPT-4 feedback~\citep{mtbench} by asking GPT-4 to compare model-generated responses with the ground truth answers\footnote{We use the ChatGPT's generation~\citep{chatgpt} for VicunaEval's ground truth responses.} and raise 1-10 scores for both responses (see Appendix \ref{app:eval_detail} for the prompt we use). We report the ratio of the total score of model responses and ground truth answers. This metric is only applied to DollyEval, SelfInst, and VicunaEval.
    \item \textbf{Human Evaluation}: We conduct human evaluations on the SelfInst dataset following ~\cite{ITGPT4} by asking volunteers to compare two responses produced by different models and annotate ``Win'', ``Tie'', or ``Loss''. More human evaluation details can be found in Appendix \ref{app:human_eval}.
\end{itemize}
For all test sets, we sample the responses with the temperature = 1 and report the average scores of 5 generations for each prompt with different random seeds.

\paragraph{Baselines} 
We consider three baselines in our main experiment:
\begin{itemize}[leftmargin=12pt,topsep=-1pt]
    \item \textbf{SFT w/o KD} directly fine-tunes the student model on $\mathcal{D}$ supervised by the golden responses.
    \item \textbf{KD}~\citep{distilbert,lightpaff} fine-tunes the student model on $\mathcal{D}$ using the teacher distribution as the supervision at each token step, also known as word-level KD.
    \item \textbf{SeqKD}~\citep{skd,vicuna,alpaca,ITGPT4,lima} fine-tunes the student model on the data generated by the teacher model.
\end{itemize}

\subsection{Results}

\begin{table}[t]
    \centering
    \small
    \begin{tabular}{lrl|cc|cc|cc|c|c}
    \toprule
    \multirow{2}{*}{Model} & \multirow{2}{*}{\#Params} & \multirow{2}{*}{Method}
    & \multicolumn{2}{c|}{DollyEval} & \multicolumn{2}{c|}{SelfInst} & \multicolumn{2}{c|}{VicunaEval} & S-NI & UnNI\\
     & & & GPT4 & R-L & GPT4 & R-L  & GPT4 & R-L & R-L & R-L \\ \midrule
     \multirow{13}{*}{GPT-2} 
     \color{gray} & 1.5B & Teacher &  58.4  &  27.6 &  42.9 &   14.3 &  48.6 & 16.3 & 27.6 & 31.8    \\ \cmidrule(l){2-11}
     & \multirow{4}{*}{120M} 
        & SFT w/o KD     &  38.6  &  23.3 &  26.3 &   10.0 &  32.8 & 14.7 & 16.3 & 18.5     \\
     &  & KD &  40.3  &  22.8 &  27.8 &   10.8 &  31.9 & 13.4 & 19.7 & 22.0     \\
     &  & SeqKD  &  41.2  &  22.7 &  26.2 &   10.1 &  31.0 & 14.3 & 16.4 & 18.8     \\
     &  & \textsc{MiniLLM}  &  \textbf{44.7}  &  \textbf{24.6} &  \textbf{29.2} &   \textbf{13.2} &  \textbf{34.1} & \textbf{16.9}* & \textbf{25.3} & \textbf{26.6}     \\ \cmidrule(l){2-11}
     & \multirow{4}{*}{340M} 
        & SFT w/o KD     &  51.9  &  \textbf{25.5}&  39.6  &  13.0         &  42.3       & 16.0    &  25.1 & 32.0    \\
     &  & KD &  51.6  &  25.0 &  39.2 &  12.0  &  42.8       & 15.4    &  23.7 & 31.0  \\
     &  & SeqKD  &  50.5  &  25.3 &  39.0 &  12.6  &  \textbf{43.0} & 16.9* & 22.9 & 30.2     \\
     &  & \textsc{MiniLLM}  &  \textbf{52.2}  &  25.4        &  \textbf{40.5}       &  \textbf{15.6}         &  42.6       & \textbf{17.7}*    &  \textbf{27.4} & \textbf{34.5}  \\ \cmidrule(l){2-11}
     & \multirow{4}{*}{760M} 
        & SFT w/o KD     &  50.7  &  25.4 &  38.3 &   12.4 &  43.1 & 16.1 & 21.5 & 27.1   \\ 
     &  & KD &  53.4  &  25.9 &  40.4 &   13.4 &  43.4 & 16.9* & 25.3 & 31.7   \\
     &  & SeqKD  &  52.0  &  25.6 &  38.9 &   14.0 &  42.4 & 15.9 & 26.1 & 32.9    \\
     &  & \textsc{MiniLLM}  &  \textbf{54.7}  &  \textbf{26.4} &  \textbf{44.6*} &   \textbf{15.9} &  \textbf{45.7} & \textbf{18.3*} & \textbf{29.3*} & \textbf{37.7*}    \\ \midrule
     \multirow{13}{*}{OPT} 
     & 13B & Teacher  &  70.3  &  29.2 &  56.1 &   18.4 &  58.0 & 17.8 & 30.4 & 36.1   \\ \cmidrule(l){2-11}
     & \multirow{4}{*}{1.3B}
        & SFT w/o KD     &  52.6  &  26.0 &  37.7 &   11.4 &  40.5 & 15.6 & 23.1 & 28.4  \\
     &  & KD &  52.7  &  25.4 &  36.0 &   12.2 &  40.8 & 14.9 & 21.9 & 27.0  \\
     &  & SeqKD  &  51.0  &  26.1 &  36.6 &   12.7 &  42.6 & 16.6 & 21.4 & 28.2  \\
     &  & \textsc{MiniLLM}  &  \textbf{60.7}  &  \textbf{26.7} &  \textbf{47.0} &   \textbf{14.8} &  \textbf{50.6} & \textbf{17.9*} & \textbf{28.6} & \textbf{33.4}   \\ \cmidrule(l){2-11}
     & \multirow{4}{*}{2.7B} 
        & SFT w/o KD     &  55.4  &  27.1 &  38.9 &   13.9 &  44.8 & 16.6 & 24.9 & 32.3    \\
     &  & KD &  60.5  &  25.9 &  48.6 &   13.8 &  51.3 & 16.7 & 26.3 & 30.2  \\
     &  & SeqKD  &  57.6  &  27.5 &  40.5 &   13.3 &  44.5 & 16.5 & 25.3 & 32.3    \\
     &  & \textsc{MiniLLM}  &  \textbf{63.2}  &  \textbf{27.4} &  \textbf{52.7} &   \textbf{17.2} &  \textbf{55.9} & \textbf{19.1*} & \textbf{30.7*} & \textbf{35.1}  \\ \cmidrule(l){2-11}
     & \multirow{4}{*}{6.7B} 
        & SFT w/o KD     &  67.9  &  27.6 &  56.4 &  16.4 &   57.3 &  17.8 & 30.3 & 28.6    \\ 
     &  & KD &  68.6  &  28.3 &  58.0 &  17.0 &   57.0 &  17.5 & 30.7* & 26.7    \\
     &  & SeqKD  &  69.6  &  28.5 &  54.0 &  17.0 &   57.6 &  17.9* & 30.4 & 28.2    \\
     &  & \textsc{MiniLLM}  &  \textbf{70.8*}  &  \textbf{29.0} &  \textbf{58.5*} &  \textbf{17.5} &   \textbf{60.1*} &  \textbf{18.7*} & \textbf{32.5*} & \textbf{36.7*}    \\ \midrule
     \multirow{5}{*}{LLaMA}
     & 13B & Teacher  & 79.0   &  29.7 &  75.5 &   23.4 &  65.1 & 19.4 & 35.8 & 38.5    \\ \cmidrule(l){2-11}
     & \multirow{4}{*}{7B}
        & SFT w/o KD     & 73.0   &  26.3 &  69.2 &   20.8 &  61.6 & 17.5 & 32.4 & 35.8   \\
     &  & KD & 73.7   &  27.4 &  70.5 &   20.2 &  62.7 & 18.4 & 33.7 & 37.9   \\
     &  & SeqKD  & 73.6   &  27.5 &  71.5 &   20.8 &  62.6 & 18.1 & 33.7 & 37.6    \\
     &  & \textsc{MiniLLM}  & \textbf{76.4}   &  \textbf{29.0} &  \textbf{73.1} &   \textbf{23.2} &  \textbf{64.1} & \textbf{20.7*} & \textbf{35.5} & \textbf{40.2*}   \\
     \bottomrule
    \end{tabular}
    \vspace{0.4cm}
    \caption{Evaluation results. GPT4 and R-L stand for the average GPT-4 feedback scores and Rouge-L scores across 5 random seeds, respectively. The best scores of each model size are \textbf{boldfaced}, and the scores where the student model outperforms the teacher are marked with *.}
    \vspace{-0.5cm}
    \label{tab:auto}
\end{table}

\label{sec:auto}
We present the R-L and GPT4 evaluation results in Table \ref{tab:auto}, from which we have three observations.

\textit{First}, by comparing the overall performance of \textsc{MiniLLM} with the baselines, we observe that the model distilled by our KD method outperforms the baselines in almost all cases, when trained with different
base models, tested on various evaluation sets, and scored by both Rouge-L and GPT-4 feedback. This verifies the good generalization and high overall performance of our KD method. We also find that \textsc{MiniLLM} generally works much better on datasets other than Dolly compared with the baselines, indicating its good out-of-distribution generalization.

\textit{Second}, the Rouge-L scores show that \textsc{MiniLLM} produces the most precise responses that have high overlaps with the ground-truth responses. In some cases, especially on Vicuna, S-NI, and UnNI, student models reach even higher Rouge-L scores than the teacher models, which matches the observation in~\cite{born_again}. We conjecture that the standard teacher-forcing fine-tuning on $\mathcal{D}$ brings training-inference discrepancy to the teacher model, also known as exposure bias~\citep{exposure_bias}. On the contrary, \textsc{MiniLLM} is optimized with policy optimization methods, which samples responses from student models during training and thus alleviates exposure bias~\citep{cold}. We include further analysis on exposure bias in Section \ref{sec:ana}.

\begin{wrapfigure}{r}{5.2cm}
\vspace{-0.8cm}
\includegraphics[width=0.38\textwidth]{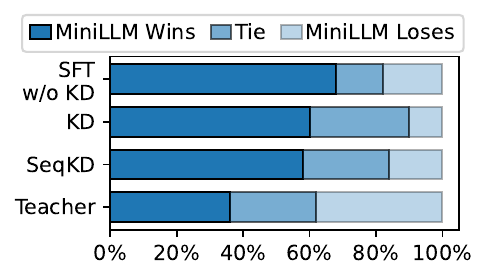}
\centering
\vspace{-0.6cm}
\caption{Human evaluation results. We use LLaMA-7B as the student and LLaMA-13B as the teacher.}
\vspace{-0.7cm}
\label{fig:human}
\end{wrapfigure}

\textit{Third}, comparing the results across model sizes and model families, we can see that the improvement of \textsc{MiniLLM} is consistent when the base model sizes vary from 120M to 13B across three model families. This tendency is also illustrated in Figure \ref{fig:full_res}, which demonstrates the excellent scalability and generalization of our method in the era of LLMs.

The human evaluation results on the SelfInst dataset based on the LLaMA family are shown in Figure \ref{fig:human}. \textsc{MiniLLM} obtains better human preference than all the baselines, performing comparably to the teacher model.

\subsection{Analysis}
\label{sec:ana}

\paragraph{Scaling Law of Teacher}
\begin{wrapfigure}{r}{5.5cm}
\vspace{-0.3cm}
\includegraphics[width=0.35\textwidth]{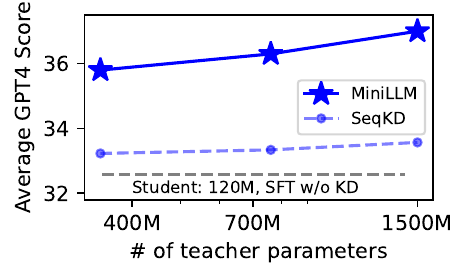}
\centering
\vspace{-0.2cm}
\caption{The scaling law of teacher based on the GPT-2 family models. We compare \textsc{MiniLLM} and SeqKD with GPT-2-125M as the student and GPT-2 340M, 760M, and 1.5B as teachers.}
\label{fig:ts_gpt2}
\end{wrapfigure}

Although it is intuitive that we can distill better student models from larger teacher models, \cite{teaching_assistant} has shown that increasing the teacher models' sizes does not guarantee the improvement of student models, sometimes even harming the distillation performance. It is not clear how \textsc{MiniLLM} works when we scale up the teacher models' sizes. Therefore, we compare \textsc{MiniLLM} and SeqKD using teacher models with different sizes and fix the size of the student model. We present the results based on the GPT-2 family in Figure \ref{fig:ts_gpt2} and that based on the OPT family in Appendix \ref{app:ts_opt}. We can see that \textsc{MiniLLM} constantly outperforms SeqKD, and the student model performance is positively correlated with the teacher model sizes. This shows the potential of our method to compress models with massive parameters.

\paragraph{Exposure Bias}


Language generation models trained to minimize \textit{forward} KLD suffer from exposure bias~\citep{exposure_bias} caused by the discrepancy between teacher-forcing training and free-run generation. When training \textsc{MiniLLM}, the student model sees samples generated by itself, alleviating the training-inference mismatch~\citep{cold}. In Figure \ref{fig:exposure_bias}, we use the ExAccErr metric~\citep{eb_measure} defined in Appendix \ref{app:eb} to measure the excess accumulated error due to exposure bias. The experiment is based on GPT-2-125M, with GPT-2-1.5B as the teacher, using Dolly as the test set. For each prompt, we sample 10 responses to reduce the variance. We can see that the ExAccErrs of the baselines continuously grow during generation, while \textsc{MiniLLM} has a much lower ExAccErr, and the error stops accumulating in long-text generation ($>$ 150 tokens).

\begin{figure}[t]
    \begin{minipage}[h]{0.50\textwidth}
        \centering
        \includegraphics[width=0.9\textwidth]{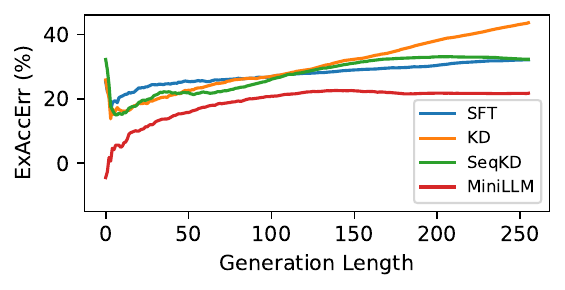}
        \caption{The excess error caused by the training-decoding discrepancy (ExAccErr) accumulated with the generation length. Lower ExAccErr means the method introduces less exposure bias.}
        \label{fig:exposure_bias}
    \end{minipage}\hspace{3mm}
    \begin{minipage}[h]{0.46\textwidth}
        \centering
        \small
        \begin{tabular}{l|cccccccc}
        \toprule
                &  \multicolumn{2}{c}{SST2} & \multicolumn{2}{c}{BoolQ} \\
                &  ECE      &    Acc.       &       ECE     &   Acc.     \\ \midrule
        Teacher &  0.025    &       93.0    &      0.356    &  74.5  \\ \midrule
        KD  &  0.191    &       84.7            &      0.682    &  63.5  \\
        SeqKD   &  0.243    &       66.5            &      0.681    &  62.8  \\
        \textsc{MiniLLM}
                &  \textbf{0.099} &  \textbf{89.7}     &      \textbf{0.502}    &  \textbf{67.8}  \\    
        \bottomrule
        \end{tabular}
        \vspace{0.3cm}\makeatletter\def\@captype{table}\makeatother\caption{The ECE and accuracy scores on SST2 and BoolQ datasets. The best scores among student models are \textbf{boldfaced}.}
        \label{tab:calibration}
    \end{minipage}

\end{figure}


\paragraph{Calibration}

\cite{gpt4} has shown that models trained with policy optimization are likely to be poorly calibrated. We test the calibration of \textsc{MiniLLM} and the KD baselines on two widely-used text classification datasets: SST2~\citep{sst-2} and BoolQ~\citep{boolq}, based on LLaMA-7B. We design zero-shot classification instructions (see Appendix \ref{app:eval_detail}) and take the probability of the label words to compute the ECE scores~\citep{ece}. From Table \ref{tab:calibration}, we observe that KD and SeqKD models are worse calibrated than the teacher model, which potentially explains their low performance on canonical benchmarks~\citep{false_imitate}. We suspect that minimizing \textit{forward} KLD causes the models to push high probabilities to void regions of the target distribution, which leads to significant distribution differences between the student and the teacher (see the example in Figure \ref{fig:toy}). In contrast, \textsc{MiniLLM} focuses on accurately learning the major parts of the target distribution and narrows the ECE scores gap between the student and the teacher.

\begin{figure}[t]
    \begin{minipage}[h]{0.50\textwidth}
        \centering
        \includegraphics[width=0.9\textwidth]{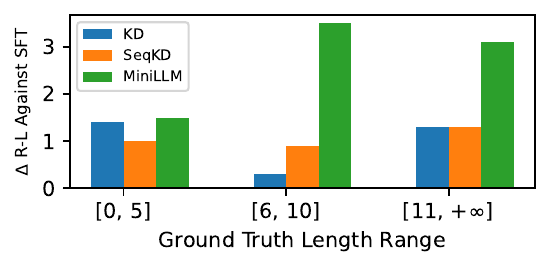}
        \makeatletter\def\@captype{figure}\makeatother\caption{The Rouge-L scores of the distilled models against SFT on the different subsets of S-NI split by the golden responses' length.}
        \label{fig:length_sinst}
    \end{minipage}\hspace{3mm}
    \begin{minipage}[h]{0.46\textwidth}
        \vspace{0.2cm}
        \centering
        \small
        \begin{tabular}{l|cccc}
        \toprule
                    & \multicolumn{2}{c}{DollyEval} & \multicolumn{2}{c}{SelfInst} \\
                    &  Dist-4 & Loss                &  Dist-4   &  Loss   \\ \midrule
        Teacher     &  99.3       &  3.55          &  99.1    &  4.44    \\
        SFT          &  99.5       &  3.89          &  99.0    &  5.28    \\
        \textsc{MiniLLM} & 99.0     &  3.95          &  98.6    &  5.33    \\ \bottomrule
        \end{tabular}
        \vspace{0.2cm}
        \makeatletter\def\@captype{table}\makeatother\caption{The distinct 4-grams (Dist-4) and language modeling loss (Loss) on the test sets based on the LLaMA family. \textsc{MiniLLM} preserves generation diversity.}
        \label{tab:diversity}
    \end{minipage}

    \vspace{-0.3cm}
\end{figure}

\paragraph{Performance on Different Response Length} We study the models' performance when the golden response lengths belong to different ranges. In Figure \ref{fig:length_sinst}, we illustrate the Rouge-L scores of different KD models against the SFT models on three S-NI subsets split by the length of the ground truth responses. We can see that all methods achieve low scores on prompts that expect short responses ($\le 5$ tokens), probably because most responses in our training set are long sentences, introducing a distribution shift between training and evaluation~\citep{ITGPT4}. Furthermore, the output spaces of these prompts are relatively small, allowing the student model to cover most modes of the teacher, and thus \textit{reverse} KLD and \textit{forward} KLD have similar performance. For prompts with longer responses ($\ge 6$ tokens), the teacher distribution contains more modes than the students due to the complex output spaces, which shows the advantage of \textsc{MiniLLM} against standard KD models. Similar results on UnNI are shown in Appendix \ref{app:length_uni}.

\paragraph{Generation Diversity} \cite{gan_falling_short} has found that the model optimized by minimizing \textit{reverse} KLD is likely to lose modes, which affects the generation diversity. We follow~\cite{cold} to discuss generation diversity from three aspects:
(i) generating multiple distinct responses given a prompt. (ii) generating linguistically complex responses. (iii) the ability to generate contents that have high coverage of the real data distribution. For (i), we argue that for many NLP applications, generating one \textbf{correct} response is sufficient, especially for those scenarios demanding high truthfulness and reliability~\citep{halu_survey}. For (ii) and (iii), we report the responses' distinct 4-gram proportion and the language modeling loss on the test sets in Table \ref{tab:diversity}, using the base models from the LLaMA family (see Appendix \ref{app:diverse} for more details) . We can see that \textsc{MiniLLM} preserves the distinct 4-gram proportion in the generated responses and language modeling loss on the test set.

\subsection{Ablation Studies on Optimization Strategies}

We evaluate the effectiveness of the three strategies proposed to stabilize and accelerate optimization in Section~\ref{sec:rl_optim} by distilling a GPT-2-125M model from the GPT-2-1.5B model. More ablation studies can be found in Appendix \ref{app:abl}. In Table \ref{tab:ablation_eval}, we report the best Rouge-L scores on the validation set of each run and the evaluation results of the corresponding checkpoints. We also plot the \textit{reverse} KLD between the student and the teacher during training in Figure \ref{fig:ablation_train}, where the curves are smoothed by 32 steps. We can see that Teacher-Mixed Sampling and Length Normalization works for stabilizing training. Although the \textit{reverse} KLDs also decrease without these strategies, we find that the models quickly learn to generate repeated, short, or meaningless strings that have high probabilities in the teacher distribution (see examples in Appendix \ref{app:cases}), which is known as reward hacking~\citep{reward_hacking}. This also leads to the low generation performance in Table \ref{tab:ablation_eval}. From Figure \ref{fig:ablation_train}, we also observe that the Single-Step Decomposition effectively reduces the variance of the training process, which also results in higher scores on the validation and test sets.

\begin{figure}[t]
    \begin{minipage}[h]{0.48\textwidth}
        \centering
        \vspace{0mm}
        \small
        \begin{tabular}{l|cc}
        \toprule
                            &  Valid. & Dolly\\
                            &    R-L        & R-L     \\ \midrule
        \textsc{MiniLLM}      & \textbf{27.4} & \textbf{24.6}   \\
        \ \ w/o Length Norm.     &  17.4        & 14.7       \\
        \ \ w/o Teacher-Mixed   &  22.3        & 20.4           \\
        \ \ w/o Single-Step   &   27.0        & 23.7         \\
        \bottomrule
        \end{tabular}

        \vspace{5mm}
        \makeatletter\def\@captype{table}\makeatother\caption{The performance on the validation and test set when different combinations of \textsc{MiniLLM} optimization strategies are applied.}
        \label{tab:ablation_eval}   
    \end{minipage}\hspace{2mm}
    \begin{minipage}[h]{0.49\textwidth}
        \centering
        \vspace{-3mm}
        \includegraphics[width=0.8\textwidth]{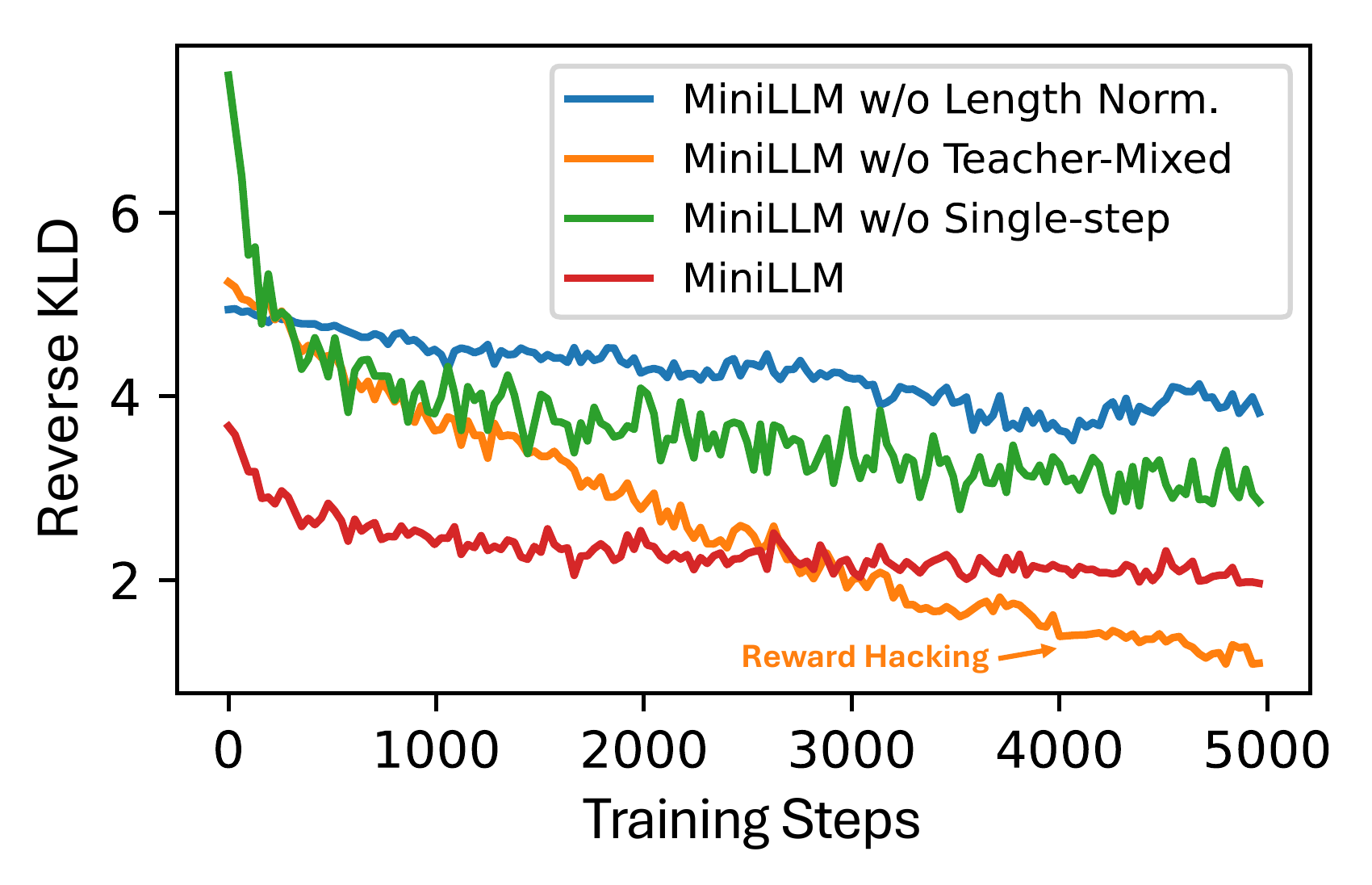}
        \vspace{-3mm}
        \makeatletter\def\@captype{figure}\makeatother\caption{The \textit{reverse} KLD between the teacher and the students during 
\textsc{MiniLLM} training when different optimization strategies are applied.}
        \label{fig:ablation_train}
    \end{minipage}
    \vspace{-0.1cm}
\end{figure}

\section{Related Work}

\paragraph{Large Language Models} Large language models (LLMs;~\citealp{gpt3,lamda,palm,gpt4,palm2}) have shown superior performance by solving various NLP tasks in a generative manner. Recent works apply instruction tuning~\citep{flan,t0,flan-t5} or learning from human feedback~\citep{instruct-gpt,hhh} to improve the alignment of LLMs with humans further and create general AI assistants~\citep{chatgpt,bard}. There are also efforts to build open-source LLMs~\citep{opt,llama,pythia} to facilitate research and industry development. Although appealing, the broad capacities of LLMs usually emerge with large model sizes~\citep{scaling_law,emergent} that require massive computational resources. Therefore, model compression is critical for the practical deployment and further research of LLMs.

\paragraph{Knowledge Distillation} Knowledge distillation (KD;~\citealp{kd}), as a widely used model compression technique, aims at training a student model with the guidance of a teacher model~\citep{policy_kd,distilbert,kdsurvey}. In the NLP community, many works apply KD to text classification tasks by mimicking the teacher model's output distribution~\citep{lightpaff,mixkd,perturb_kd}, hidden states~\citep{tinybert,bert-pkd}, or attention scores~\citep{minilm,minilmv2}. For text generation, the standard KD method is to approximately minimize the \textit{forward} KLD between the student's and the teacher's generation distribution by using the teacher's output at each time step as supervision~\citep{distilbert} or direct training on the teacher's generated texts~\citep{skd,alpaca,vicuna,ITGPT4}. In this paper, we minimize the \textit{reverse} KLD, which is more suitable for LLMs when the teacher distribution is available. Concurrent works~\citep{gkd,f-div-kd} also explore more the distribution discrepancy metrics in KD.

\paragraph{Distribution Discrepancy Metrics in Text Generation} The distribution discrepancy metrics play a significant role in training text generation models. The \textit{forward} KLD is widely used due to its simplicity when derived as the Maximum Likelihood Estimate (MLE) objective~\citep{zhang2018minimum}. However, previous works show that minimizing \textit{forward} KLD leads to zero-forcing behavior where models try to cover all modes of the target distribution and sacrifice the accuracy of major modes~\citep{huszar2015not}. Some works resort to using other metrics to remedy this problem, such as \textit{reverse} KLD~\citep{smart}, Total Variation Distance~\citep{tailor}, and Optimal Transport~\citep{ot_for_gen}. Our paper tackles this problem under the scenario of knowledge distillation for LLMs.

\section{Conclusion}

In this work, we investigate the problem of distilling the knowledge of LLMs into small language models. We find that the standard distillation methods minimizing the \textit{forward} KLD is sub-optimal in language generation scenarios because the teacher's output distribution contains more modes than the student's, and \textit{forward} KLD forces the student distribution to overestimate the low-probability regions of the teacher distribution. Therefore, we propose \textsc{MiniLLM} that minimizes the $\textit{reverse}$ KLD between the teacher and student distribution and design an algorithm to optimize this objective. Extensive experiments show that \textsc{MiniLLM} produce more precise responses that have higher overall quality than standard KD models. We also find that \textsc{MiniLLM} has lower exposure bias, better calibration, and higher performance in long-text generation with good generation diversity. 

\section*{Acknowledgements}
This work was supported by the National Key Research and Development Program of China (No. 2021ZD0113304), the National Science Foundation for Distinguished Young Scholars (with No. 62125604), and the NSFC projects (Key project with No. 61936010). 

\bibliographystyle{alpha}
\bibliography{neurips_2023}

\newpage
\appendix
\section{Derivations}

\subsection{A Perspective of \textsc{MiniLLM} from Inverse Reinforcement Learning}
\label{app:persp_rl}
In Section \ref{sec:minillm}, we formulate KD as an optimization problem of minimizing the discrepancy between the teacher distribution and the student distribution and finally reach the objective of minimizing \textit{reverse} KLD.
Alternatively, we can also regard KD as training the student model with the teacher model's guidance, which resembles an agent learning from the feedback from an environment. Following \cite{cold}, we treat token generation as a Markov Decision Process. At each time step $t$, the student model chooses an action (token) $y_t$ from the action space (vocabulary) $V$ conditioning on the state (prefix) $(\by_{<t},\bx)$ based on the policy (generation probability) $\pst$. 

From this perspective, standard KD corresponds to behavior cloning (BC;~\citealp{behavior_cloning}) in imitation learning~\citep{imitation_learning}. However, BC is known to under-perform Inverse Reinforcement Learning (IRL;~\citealp{maxent_irl}), another imitation learning method that first recovers a reward model from the environment demonstrations and then trains the policy with the reward using policy optimization algorithms~\citep{policy_gradient,ppo}. Therefore, in the KD scenario, we seek to first induce a reward $r(y_t, (\by_{<t},\bx))$ from the environment (the teacher model) and then train the student model to maximize the reward as the objective. We take the maximum-entropy inverse reinforcement learning framework~\citep{maxent_irl,scalable_irl} and thus the
Q-function $Q(y_t, (\by_{<t},\bx))$ in the environment satisfies the soft Bellman Equation:
\begin{equation}
    \small
    Q(y_t, (\by_{<t},\bx)) = r(y_t, (\by_{<t},\bx)) + \gamma \log \sum_{y'\in V} \operatorname{exp}[Q(y', (\by_{\le t}, \bx))].
\end{equation}
We follow \cite{irl_gen} to parameterize the Q-function as $Q(y_t, (\by_{<t}, \bx)) = f(y_t, (\by_{<t}, \bx))$ and assume $\gamma=1$, where $f(y_t, (\by_{<t}, \bx))$ is the output logits of the teacher model\footnote{The teacher model's distribution satisifies $\ptt=\frac{\operatorname{exp}(f(y_t, (\by_{<t},\bx)))}{\sum_{y'\in V}\operatorname{exp}(f(y', (\by_{<t},\bx)))}$.}. Then, the reward is given by:
\begin{equation}
    \small
    r(y_t, (\by_{<t},\bx)) = f(y_t, (\by_{<t},\bx)) - \log \sum_{y'\in V} \operatorname{exp}[f(y', (\by_{\le t}, \bx))].
\end{equation}
To maximize the reward, we apply maximum-entropy reinforcement learning~\cite{max_ent_rl}, whose learning objective is 
\begin{equation}
    \label{eq:rl_obj}
    \small
    \max_{\theta} \mathcal{J}(\theta) = \max_{\theta} \expct_{\substack{\bx \sim p_{\bx} \\ \by \sim \ps (\cdot | \bx)}} \sum_{t=1}^{|\by|} \left[r(y_t, (\by_{<t},\bx)) + \operatorname{H}\left[q_{\theta}(\cdot | \by_{<t}, \bx)\right]\right],
\end{equation}
where $\operatorname{H}\left[q_{\theta}(\cdot | \by_{<t}, \bx)\right] = -\expct_{y_t\sim q_{\theta}(\cdot | \by_{<t}, \bx)}\log q_{\theta}(\cdot | \by_{<t}, \bx)$ is the entropy of the student model distribution at the time step $t$. 
\vspace{-0.1cm}
\paragraph{Equivalence Between Objectives} We prove an approximate equivalence between Eq. \ref{eq:rl_obj} and Eq. \ref{eq:obj}. We first rewrite the summation of the reward $\sum_{t=1}^{|\by|} r(y_t, (\by_{<t},\bx))$ by the associative law:
{\small
\begin{align}
    \sum_{t=1}^{|\by|} r(y_t, (\by_{<t},\bx)) =& \sum_{t=1}^{|\by|} \left[f(y_t, (\by_{<t},\bx)) - \log \sum_{y'\in V} \operatorname{exp}[f(y', (\by_{\le t}, \bx))]\right] \\
    =& \ f(y_1, (\by_{<1},\bx)) + \sum_{t=2}^{|\by|} \left[f(y_t, (\by_{<t},\bx)) - \log \sum_{y'\in V} \operatorname{exp}[f(y', (\by_{\color{red}{<t}}, \bx))]\right] \\
    & - \log \sum_{y'\in V} \operatorname{exp}[f(y', (\by_{\le |\by|}, \bx))] \\
    \approx & \sum_{t=1}^{|\by|} \left[f(y_t, (\by_{<t},\bx)) - \log \sum_{y'\in V} \operatorname{exp}[f(y', (\by_{\color{red}{<t}}, \bx))]\right] \\
    =& \sum_{t=1}^{|\by|} \log \frac{\operatorname{exp}(f(y_t, (\by_{<t},\bx)))}{\sum_{y'\in V}\operatorname{exp}(f(y', (\by_{<t},\bx)))} \\
    =& \sum_{t=1}^{|\by|} \log \ptt.
\end{align}
}
Then, $\mathcal{J}(\theta)$ can be approximately rewritten as:
{\small
\begin{align}
    \mathcal{J}(\theta) \approx &\expct_{\substack{\bx \sim p_{\bx} \\ \by \sim \ps (\cdot | \bx)}} \sum_{t=1}^{|\by|} \left[\log \ptt + \operatorname{H}\left[q_{\theta}(\cdot | \by_{<t}, \bx)\right]\right] \\
    = & \expct_{\substack{\bx \sim p_{\bx} \\ \by \sim \ps (\cdot | \bx)}}  \sum_{t=1}^{|\by|} \left[\log \ptt -\log \left[q_{\theta}(\cdot | \by_{<t}, \bx)\right]\right] \\
    = & -\operatorname{KL}(\ps || \pt) \\
    = & -\mathcal{L}(\theta).
\end{align}
}
Therefore, maximizing $\mathcal{J}(\theta)$ is approximately equivalent to minimizing $\mathcal{L}(\theta)$.

\subsection{Derivation of Equation \ref{eq:grad}}
\label{app:grad}
We compute the gradient of $\mathcal{L}(\theta)=\revkl$ with respect to $\theta$ using the Policy Gradient Theorem~\citep{policy_gradient}:
\begin{equation}
\small
\begin{aligned}
    \nabla \mathcal{L}(\theta) 
    &= -\nabla \expct_{\substack{\bx \sim p_{\bx} \\ \by \sim \ps (\cdot | \bx)}} \R \\
    &= -\int \nabla \left[\ps (\by | \bx) \R \right] \dy \dx\\
    &= -\int \ps(\by | \bx) \nabla \R \dy \dx - \int \R \nabla \ps(\by|\bx) \dy \dx \\
    &= \int \ps(\by | \bx) \nabla \log \ps(\by|\bx)\dy \dx - \int \ps(\by|\bx) \R \nabla \log \ps(\by|\bx) \dy \dx \\
    &= -\expct_{\substack{\bx \sim p_{\bx} \\ \by \sim \ps (\cdot | \bx)}} (\R - 1) \nabla \log \ps (\by | \bx) \\
    &= -\expct_{\substack{\bx \sim p_{\bx} \\ \by \sim \ps (\cdot | \bx)}} \sum_{t=1}^T (\sum_{t'=1}^T \log \frac{\pttp}{\pstp} - 1) \nabla \log \pst \\
    &= -\expct_{\substack{\bx \sim p_{\bx} \\ \by \sim \ps (\cdot | \bx)}} \sum_{t=1}^T (\sum^T_{t'=\color{red}{t}} \log \frac{\pttp}{\pstp} - 1) \nabla \log \pst, \label{eq:app_grad}
\end{aligned}
\end{equation}
where Eq. \ref{eq:app_grad} is based on the fact that $\log \pst$ can only affect tokens at $\ge t$ positions in $\by$. By setting $R_t =  \sum_{t'=t}^T \log \frac{\pttp}{\pstp}$, we obtain Eq. \ref{eq:grad}.

\subsection{Derivation of Equation \ref{eq:exp_reg}}
\label{app:exp_reg}

To derive Eq. \ref{eq:exp_reg}, we first denote: 
\begin{equation}
    \small
    \begin{aligned}
    (\nabla \mathcal{L})_{\text{Single}}=&-\expct_{\substack{\bx \sim p_{\bx} \\ \by \sim \ps (\cdot | \bx)}} \left[ \sum_{t=1}^T\nabla \expct_{y_t \sim q_\theta (t)}[r_t]\right], \\
    (\nabla \mathcal{L})_{\text{Long}} =& -\expct_{\substack{\bx \sim p_{\bx} \\ \by \sim \ps (\cdot | \bx)}} \sum_{t=1}^T R_{t+1} \nabla \log \pst.
\end{aligned}
\end{equation}
Then, we re-write $\nabla \mathcal{L}(\theta)$ as:
{\small
\begin{align}
    \nabla \mathcal{L}(\theta) 
    =& -\expct_{\substack{\bx \sim p_{\bx} \\ \by \sim \ps (\cdot | \bx)}} \sum_{t=1}^T (R_t - 1) \nabla \log \pst \\
    =& -\expct_{\substack{\bx \sim p_{\bx} \\ \by \sim \ps (\cdot | \bx)}} \sum_{t=1}^T R_{t+1} \nabla \log \pst \\
    &- \expct_{\substack{\bx \sim p_{\bx} \\ \by \sim \ps (\cdot | \bx)}} \sum_{t=1}^T \left(\log \frac{\ptt}{\pst}-1 \right) \nabla \log \pst \\
    =& (\nabla \mathcal{L})_{\text{Long}} - \expct_{\substack{\bx \sim p_{\bx} \\ \by \sim \ps (\cdot | \bx)}} \sum_{t=1}^T \expct_{y_t \sim \ps (\cdot | \by_{<t}, \bx)} \left(\log \frac{\ptt}{\pst}-1 \right) \nabla \log \pst \\ \label{eq:app_kl}
    =& (\nabla \mathcal{L})_{\text{Long}} - \expct_{\substack{\bx \sim p_{\bx} \\ \by \sim \ps (\cdot | \bx)}} \sum_{t=1}^T \nabla \expct_{y_t \sim \ps (\cdot | \by_{<t}, \bx)} \left[- \log \frac{\pst}{\ptt}\right] \\ 
    =& (\nabla \mathcal{L})_{\text{Long}} - \expct_{\substack{\bx \sim p_{\bx} \\ \by \sim \ps (\cdot | \bx)}} \left[ \sum_{t=1}^T\nabla \expct_{y_t \sim q_\theta (t)}[r_t]\right] \\
    =& (\nabla \mathcal{L})_{\text{Long}} + (\nabla \mathcal{L})_{\text{Single}},
\end{align}
}%
where Eq. \ref{eq:app_kl} uses the product rule of the gradient and $r_t = \log \frac{\ptt}{\pst}$.

\section{Experimental Details}

\subsection{Training Details}
\label{app:training_detail}
\paragraph{Baselines} Our baselines include \textbf{SFT w/o KD}, \textbf{KD}, and \textbf{SeqKD}. For models with less than 1.3B parameters, we search for the learning rates in [5e-4, 1e-4, 5e-5], the batch sizes in [32, 64], and train these models for 20 epochs. For other models, we search for the learning rate in [5e-5, 1e-5, 5e-6], the batch sizes in [32, 64], and train these models for 10 epochs.
For \textbf{KD}, we follow \cite{lightpaff} to mix the distillation loss with the language modeling loss on the ground truth responses by a mixture rate of 0.5. The checkpoints of each baseline are selected by the Rouge-L~\citep{rouge} scores on the validation set because, as stated in previous works~\citep{super-natural-instructions,instruct-gpt}, we also find that Rouge-L is better correlated with human judgments.

\paragraph{\textsc{MiniLLM}} As stated in Section \ref{sec:alg}, training of \textsc{MiniLLM} has two phases which is similar to Reinforcement Learning from Human Feedback (RLHF;\citealp{instruct-gpt}).
\begin{itemize}[leftmargin=12pt,noitemsep,topsep=-3pt]
    \item \textbf{Phase 1}: We fine-tune the student model on the instruction-response training set $\mathcal{D}$ to get a starting point for the subsequent \textsc{MiniLLM} training. We fine-tune the model for 3 epochs using the best learning rate and batch size of the corresponding \textbf{SFT w/o KD} baselines. Note that different from the \textbf{SFT w/o KD} baseline, we select the checkpoint with the \textit{lowest validation loss}, not the Rouge-L score in this phase.
    \item \textbf{Phase 2}: We continuously train the model from \textbf{Phase 1} as described in Algorithm \ref{alg:minillm} using a learning rate 5e-6, a mini-batch size $64$ in all cases. The training and validation set are same as in \textbf{Phase 1}. Similar to~\cite{instruct-gpt}, we collect 256 sentences at once and adopt 4 inner epochs when doing the policy optimization. The clipping rate $\epsilon$ is set to 0.2, and the max length of the model is 512. We use temperature = 1 when sampling from $\ps$. We train the model for 5000 steps and select the final checkpoint using the Rouge-L score on the validation set. Our experiments are based on the NVIDIA V100 32G GPUs. Distilling LLaMA-7B from LLaMA-13B takes less than 10 ours on 16 GPUs.
\end{itemize}

\subsection{Automatic Evaluation Details}
\label{app:eval_detail}
During the evaluation, we sample the responses from each model using temperature = 1, a max-length limit of 512, and random seeds [10, 20, 30, 40, 50]. Similar to \cite{alpaca}, we adopt a prompt wrapper shown in Figure \ref{fig:prompt_wrapper} to convert each instruction-response pair to a sentence. For the GPT-4 feedback, we apply the prompt in Figure \ref{fig:prompt_gpt4} and set the temperature = 0.7. For the classification tasks in the ``Calibration'' paragraph of Section \ref{sec:ana}, we prompt the model to do zero-shot text classification with the templates in Figure \ref{fig:prompt_sst2} and \ref{fig:prompt_boolq}.

\begin{figure}[t]
    \begin{tcolorbox}
    Below is an instruction that describes a task. \\
    Write a response that appropriately completes the request. \\ \\
    \#\#\# Instruction: \\
    \{instruction\} \\ \\
    \#\#\# Input: \\
    \{input\} \\ \\
    \#\#\# Response:
    \end{tcolorbox}
    \caption{The prompt wrapper for training and evaluation.}
    \label{fig:prompt_wrapper}
\end{figure}

\begin{figure}
    \begin{tcolorbox}
        We would like to request your feedback on the performance of two AI assistants in response to the user instruction and input displayed above. \\
        Please rate the helpfulness, relevance, accuracy, and level of detail of their responses. Each assistant receives an overall score on a scale of 1 to 10, where a higher score indicates better overall performance. \\
        Please first output a single line containing only two values indicating the scores for Assistant 1 and 2, respectively. The two scores are separated by a space. \\
        In the subsequent line, please provide a comprehensive explanation of your evaluation, avoiding any potential bias and ensuring that the order in which the responses were presented does not affect your judgment.
    \end{tcolorbox}
    \caption{GPT-4 evaluation prompt.}
    \label{fig:prompt_gpt4}
\end{figure}

\begin{figure}
    \begin{tcolorbox}
    Below is an instruction that describes a task. \\
    Write a response that appropriately completes the request. \\ \\
    \#\#\# Instruction: \\
    Determine the sentiment of the input sentence. Please respond as positive or negative. \\ \\
    \#\#\# Input: \\
    \{sentence\} \\ \\
    \#\#\# Response:
    \end{tcolorbox}
    \caption{Zero-shot text classification prompt for SST2.}
    \label{fig:prompt_sst2}
\end{figure}

\begin{figure}
    \begin{tcolorbox}
    Below is an instruction that describes a task. \\
    Write a response that appropriately completes the request. \\ \\
    \#\#\# Instruction: \\
    Read the input passage and answer the question: \{question\}? Your answer should be ``Yes'' or ``No''. \\ \\
    \#\#\# Input: \\
    \{passage\} \\ \\
    \#\#\# Response:
    \end{tcolorbox}
    \caption{Zero-shot text classification prompt for BoolQ.}
    \label{fig:prompt_boolq}
\end{figure}

\subsection{Human Evaluation Details}
\label{app:human_eval}
Following \cite{ITGPT4}, we use SelfInst~\citep{self_inst} to perform human evaluation. We randomly sampled 50 prompts because we found that more prompts do not affect the results much. We ask the annotators to compare the responses generated by the baseline models with \textsc{MiniLLM} and decide which response is preferred or neither of them is significantly better. Note that which model the responses come from is invisible to the annotators. The interface presented to annotators is shown in Figure \ref{fig:hm_interface}.

\begin{figure}[t]
    \begin{tcolorbox}
Below is an instruction that describes a task, paired with an input that provides further context. Write a response that appropriately completes the request. \\

\#\#\# Instruction:

Desk jobs require writing a lot of emails, so it isn't surprising we get tired of repeating ourselves. Come up with several synonyms for the given word. \\

\#\#\# Input:

Sincerely \\

\#\#\# Response: \\ \\

\#\#\#\#\# Answer \#1 \#\#\#\#\# 

Fondly, affectionately, lovingly, tenderly, honestly, truly, faithfully, devotedly, passionately \\

\#\#\#\#\# Answer \#2 \#\#\#\#\# 

Faithfully, Gullibly, Humbly, Piously, Strangely, Weirdly, Yours truly \\

1: Answer \#1 is better

2: Answer \#2 is better

3: Tie

Your choice:
    \end{tcolorbox}
    \caption{The prompt wrapper for training and evaluation.}
    \label{fig:hm_interface}
\end{figure}

\subsection{Details About Generation Diversity Metrics}
\label{app:diverse}
In Table \ref{tab:diversity}, we report the distinct 4-grams (Dist-4) and the language modeling loss (Loss) on the test sets. More details about these two metrics are as follows:
\begin{itemize}[leftmargin=12pt,topsep=-1pt]
    \item \textbf{``Dist-4''} is a fraction: $N/C$ , where $N$ is the number of the distinct 4-grams in the generated responses and $C$ is the total number of 4-grams. It is a widely used metric to measure the generation diversity of a language model~\citep{dist4}. The $(N/C)$s on the Dolly test set across 5 random seeds are shown in Table \ref{tab:app_diverse}. Table \ref{tab:diversity} reports the average values across the 5 random seeds.
    \item \textbf{``Loss''} is the negative log-likelihood loss on the test set $\mathcal{D}_{\text{Test}}$: $-\sum_{\bx,\by\sim \mathcal{D}_{\text{Test}}}\log q_\theta(\by|\bx)$. It measures the mode coverage of the real data distribution because it is essentially the forward KLD between the real data distribution and the model output distribution. This relates to diversity as in the ability to generate different generations given one context with different random seeds.
\end{itemize}

\begin{table}[t]
    \centering
    \small
    \begin{tabular}{lccccc}
        \toprule
       \diagbox{Model}{Seed}         &  10 & 20 & 30 & 40 & 50\\ \midrule
       Teacher  & 23562 / 23696 & 23653 / 23834 & 24306 / 24488 & 24207 / 24381 & 23803 / 23967  \\ \midrule
       KD       & 25889 / 26064 & 24024 / 24197 & 25663 / 25843 & 25611 / 25763 & 26178 / 26339 \\ 
       SeqKD    & 25358 / 25519 & 25631 / 25822 & 26190 / 26370 & 25574 / 25748 & 26295 / 26522 \\
       \textsc{MiniLLM} & 24187 / 24458 & 25011 / 25272 & 25100 / 25436 & 24067 / 24312 & 25205 / 25519 \\ 
       \bottomrule
    \end{tabular}
    \vspace{0.5cm}
    \caption{The ($N/C$)s, where $N$ is the number of the distinct 4-grams in the generated responses and $C$ is the total number of 4-grams. We report the numbers computed on the Dolly test set when evaluated with 5 random seeds: [10, 20, 30, 40, 50].}
    \label{tab:app_diverse}
\end{table}

\subsection{Exposure Bias Analysis}
\label{app:eb}
Following \cite{eb_measure}, we compute the ExAccErr with the following formula:
\begin{align}
    \text{ExAccErr} (l)= \frac{R(l) - l\epsilon(l)}{l\epsilon(l)} \times 100\%,
\end{align}
where $R(l)$ is the accumulated regret of imitating the teacher distribution $p$ at the time step $l$ during the free-run generation:
\begin{equation}
    R(l) = \sum_{t=1}^T\expct_{\substack{\by_{<t} \sim \ps(\cdot | \bx) \\ y_t \sim \pt(\cdot | \by_{<t}, \bx) }} \log \frac{\ptt}{\pst},
\end{equation}
and $\epsilon(l)$ is the average per-step error between $\ps$ and $\pt$ using the oracle context sampled from $\pt$ as the prefix:
\begin{equation}
    \epsilon(l) = \frac{1}{l} \sum_{t=1}^T\expct_{\substack{\by_{<t} \sim p(\cdot | \bx) \\ y_t \sim \pt(\cdot | \by_{<t}, \bx) }} \log \frac{\ptt}{\pst}.
\end{equation}
Intuitively, the regret of $\ps$ during generation is made of two parts: the error to estimate $\pt$ given the oracle context and the error caused by the low-quality model-generated prefix. The former is calculated by $l\epsilon(l)$, and the latter reflects the exposure bias. Therefore, ExAccErr measures the relative error caused only by exposure bias.

\section{Additional Results}

\subsection{GPT-J as the Teacher Model}
\label{app:res_gptj}
We present the evaluation results when using GPT-J as the teacher model and GPT-2-760M, GPT-2-1.5B, and GPT-\textit{Neo}-2.7B~\citep{gpt-neo} as the student models in Table \ref{tab:auto_gptj}. \textsc{MiniLLM} outperforms the baselines in most cases.

\begin{table}[t]
    \centering
    \small
    \begin{tabular}{ll|cc|cc|cc|c|c}
    \toprule
    \multirow{2}{*}{Model} & \multirow{2}{*}{Method}
    & \multicolumn{2}{c|}{DollyEval} & \multicolumn{2}{c|}{SelfInst} & \multicolumn{2}{c|}{VicunaEval} & S-NI & UnNI\\
     & & GPT4 & R-L & GPT4 & R-L  & GPT4 & R-L & R-L & R-L \\ \midrule
     GPT-J-6B & Teacher &  65.8  &  27.3 &  57.4 &   17.3 &  55.8 & 17.4 & 28.0 & 33.6    \\ \cmidrule(l){1-10}
     \multirow{4}{*}{GPT-2-760M} 
        & SFT w/o KD     &  50.7  & 25.4 &  38.3 &   12.4 &  43.1 & 16.1 & 21.5 & 27.1     \\
     & KD &  51.6  &  \textbf{26.7} &  38.9 &   13.4 &  43.4 & 16.4 & 25.9 & 33.2     \\
     & SeqKD  &  51.4  &  26.0 &  39.2 &   14.0 &  42.0 & 15.3 & 25.5 & 32.5     \\
     & \textsc{MiniLLM}  &  \textbf{54.0}  &  25.8 &  \textbf{43.7} &   \textbf{16.3} &  \textbf{44.3} & \textbf{19.1}* & \textbf{27.1} & \textbf{35.5}*     \\ \cmidrule(l){1-10}
     \multirow{4}{*}{GPT-2-1.5B} 
        & SFT w/o KD     &  58.4  &  \textbf{27.6}* &  42.9  &  14.3         &  48.6       & 16.3    &  27.6 & 34.6*    \\
     & KD &  56.5  &  26.6 &  46.0 &  14.5  &  47.2       & 16.5    &  27.6 & 34.9*  \\
     & SeqKD  &  58.5  &  27.0 &  43.2 &  13.6  &  46.6 & 16.9 & 28.0 & 34.2*     \\
     & \textsc{MiniLLM}  &  \textbf{59.6}  &  25.9        &  \textbf{48.5}       &  \textbf{16.6}         &  \textbf{48.9}       & \textbf{19.4}*    &  \textbf{28.5}* & \textbf{35.9}*  \\ \cmidrule(l){1-10}
     \multirow{4}{*}{GPT-\textit{Neo}-2.7B} 
        & SFT w/o KD     &  60.7  &  26.8 &  45.4 &   15.8 &  51.5 & 17.0 & 26.5 & 31.6   \\ 
     & KD &  61.5  &  26.7 &  47.0 &   16.0 &  52.1 & 16.9 & 27.2 & 32.7   \\
     & SeqKD  &  60.8  &  25.6 &  47.2 &   16.2 &  53.0 & 16.9 & 26.1 & 32.9    \\
     & \textsc{MiniLLM}  &  \textbf{63.4}  &  \textbf{28.5*} &  \textbf{52.5} &   \textbf{17.1} &  \textbf{54.1} & \textbf{18.6*} & \textbf{29.8*} & \textbf{35.4*}    \\ 
     \bottomrule
    \end{tabular}
    \vspace{0.4cm}
    \caption{Evaluation results when GPT-J is the teacher. GPT4 and R-L stand for the average GPT-4 feedback scores and Rouge-L scores across 5 random seeds. The best scores of each model size are \textbf{boldfaced}, and the scores where the student model outperforms the teacher are marked with *.}
    \label{tab:auto_gptj}
\end{table}

\begin{figure}[t]
    \begin{minipage}[h]{0.50\textwidth}
        \centering
        \includegraphics[width=0.6\textwidth]{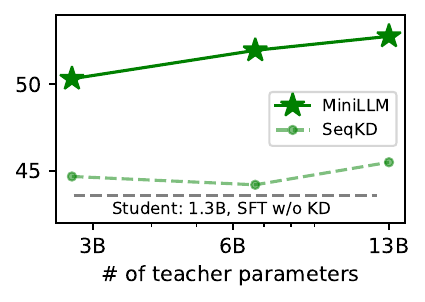}
        \caption{The scaling law of teacher model based on the OPT family models. We compare \textsc{MiniLLM} and SeqKD with OPT-1.3M as the student and OPT 2.7B, 6.7B, and 13B as teachers.}
        \label{fig:ts_opt}
    \end{minipage}\hspace{3mm}
    \begin{minipage}[h]{0.46\textwidth}
        \centering
        \includegraphics[width=0.9\linewidth]{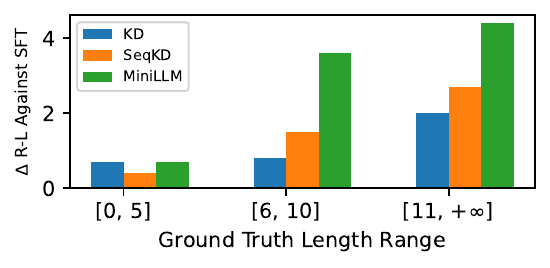}
        \caption{The Rouge-L scores of the distilled models against the SFT models on the different evaluation subsets of UnNI split by the golden responses' length.}
        \label{fig:length_uinst}
    \end{minipage}
\end{figure}

\subsection{Scaling Law of Teacher based on the OPT family} 
\label{app:ts_opt}

We present the performance of \textsc{MiniLLM} and SeqKD when we scale up the sizes of teacher models in Figure \ref{fig:ts_opt}.
Similar to the observations in Section \ref{sec:ana}, \textsc{MiniLLM} constantly performs better and distills better student models from larger teacher models.

\subsection{Performance of Response Length on U-NI}
\label{app:length_uni}
The performance on different U-NI subsets split by the length of the ground truth response is shown in Figure \ref{fig:length_uinst}.  We have the same observation as in Section \ref{sec:ana} that on short responses, all KD methods perform similarly, and on long responses, \textsc{MiniLLM} outperforms other methods.

\begin{figure}[t]
    \begin{minipage}[h]{0.48\textwidth}
        \centering
        \includegraphics[width=0.85\textwidth]{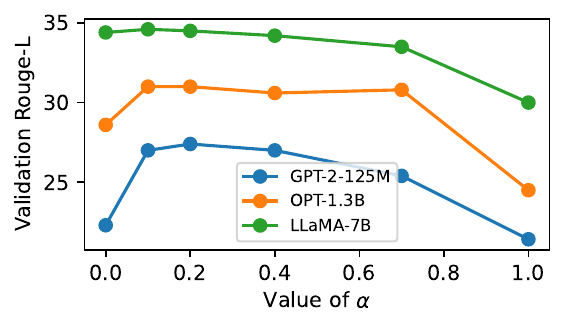}
        \vspace{-2mm}
        \makeatletter\def\@captype{figure}\makeatother\caption{Effect of the $\alpha$ value in the teacher mix-in exploration on the validation Rouge-L score. Larger models to more robust to $\alpha$.}
        \label{fig:ablation_alpha}
    \end{minipage}\hspace{2mm}
    \begin{minipage}[h]{0.48\textwidth}
        \centering
        \vspace{3mm}
        \small
        \begin{tabular}{ll|cc}
        \toprule
                        &    &  CLS  & Inst.       \\\midrule
        \multirow{2}{*}{1.3B}
            & \textsc{MiniLLM} &  \textbf{70.2}        & \textbf{52.8}  \\
            & \ \ w/o $\mathcal{L}_{\text{PT}}$  &  65.7        & 53.2  \\ \midrule
        \multirow{2}{*}{7B}
            & \textsc{MiniLLM} &  \textbf{78.8}        & \textbf{71.2}  \\ 
            & \ \ w/o $\mathcal{L}_{\text{PT}}$  &  74.3        & 71.1  \\ \bottomrule
        
        \end{tabular}
        \vspace{3mm}
        \makeatletter\def\@captype{table}\makeatother\caption{The effect of adding the pre-training loss. ``CLS'' is the average accuracy scores on SST2 and BoolQ. ``Inst.'' is the average Rouge-L score on Dolly, SelfInst, and Vicuna.}
        \label{tab:ablation_ptx}   
    \end{minipage}
\end{figure}

\subsection{More Ablation Studies}
\label{app:abl}

\paragraph{Effect of Teacher-Mix-in Strength $\alpha$} In Figure \ref{fig:ablation_alpha}, we plot the best Rouge-L scores on the validation set of GPT-2-125M, OPT-1.3B, and LLaMA-7B using GPT-2-1.5B, OPT-13B, and LLAMA-13B as the teachers, with different teacher-mix-in strength $\alpha$ in \textsc{MiniLLM}. $\alpha=0.0$ means we only sample from the student distribution, and when $\alpha=1.0$, we sample entirely from the teacher distribution. We find that $\alpha=0.2$ is generally suitable across different model families and sizes, and larger models are more robust to the choice of $\alpha$. 

\paragraph{Effect of Adding Pre-Training Loss} In Table \ref{tab:ablation_ptx}, we study the effect of adding the pre-training loss in Algorithm \ref{alg:minillm} by comparing \textsc{MiniLLM} with its variant where the language modeling loss on the pre-training corpus is removed (w/o $\mathcal{L}_{\text{PT}}$). We have a similar observation as \cite{instruct-gpt} that adding the pre-training loss helps to preserve the abilities on canonical NLP tasks while keeping the performance on instruction-following tasks nearly unchanged.

\section{Cases}
We provide some cases generated by the models distilled by different methods based on the LLaMA model family in Table \ref{tab:cases}. The prompts are sampled from the SelfInst dataset. We find that \textsc{MiniLLM} generates more detailed and accurate responses compared with the baselines.

\label{app:cases}
\begin{table}[t]
    \centering
    \begin{tabular}{l|p{10.3cm}}
    \toprule
    \multicolumn{2}{c}{Case \#1}          \\ \midrule
    Instruction     &  During writing, we added an asterisk for the word that did not come to mind. You will need to provide several examples to demonstrate all the words that can be used in the sentence instead of the asterisk.\\ \midrule
    Input           &  we * to know \\ \midrule \midrule
    SFT w/o KD      &  We can use ``know'' in the sentence instead of ``we * to know'' \\ \midrule
    KD              &  we went to know$\backslash$n we will go to know$\backslash$n we can go to know$\backslash$n we may go to know$\backslash$n we should go to know$\backslash$n we must go to know\\ \midrule
    SeqKD           &  we do to know$\backslash$n we can do to know$\backslash$n we should do to know$\backslash$n we would do to know$\backslash$n we ought to do to know \\ \midrule
    \textsc{MiniLLM}  &  We can use the following words in place of the asterisk:$\backslash$n - want$\backslash$n - have$\backslash$n - do$\backslash$n - used$\backslash$n - ought$\backslash$n - desire$\backslash$n - need$\backslash$n - must             \\ \midrule
    \makecell[l]{\textsc{MiniLLM} \\ w/o teacher-mixed} &  We to know \\
    \toprule
    \toprule
    \multicolumn{2}{c}{Case \#2}          \\ \midrule
    Instruction     &  Take the title of the product and extract its attributes. The attributes in this case refer to the characteristics of the products, such as their brand and color. Your answer must a list following the format of ``attribute: value''.\\ \midrule
    Input           &  SAMSUNG 980 PRO SSD 2TB PCIe NVMe Gen 4 Gaming M.2 Internal Solid State Hard Drive Memory Card, Maximum Speed, Thermal Control, MZ-V8P2T0B \\ \midrule \midrule
    SFT w/o KD      &  Attribute: Maximum Speed$\backslash$n Value: 3500 MB/s \\ \midrule
    KD              &  attribute: brand, attribute: color, attribute: memory card, attribute: maximum speed, attribute: thermal control\\ \midrule
    SeqKD           &  brand: Samsung, color: black, size: 2TB, interface: PCIe NVME Gen 4, design: M.2 Internal \\ \midrule
    \textsc{MiniLLM}  &  brand: SAMSUNG$\backslash$n color: 980 PRO$\backslash$n capacity: 2TB$\backslash$n interface: PCIe NVME Gen 4$\backslash$n speed: Maximum Speed$\backslash$n threshold: Thermal Control$\backslash$n form factor: M.2 Internal \\ \midrule
    \makecell[l]{\textsc{MiniLLM} \\ w/o teacher-mixed} &  SAMSUNGC 980 Pro SSD 2TB PCIe NVME Gen 4 Gaming M.II Internal Solid State Hard Drive memory Card, Max Speed, Thermal Control, AP1500K \\
    \bottomrule
    \end{tabular}
    \vspace{0.4cm}
    \caption{Instruction-following cases from the SelfInst dataset. \textsc{MiniLLM} better follows the instructions and provides more detailed and accurate responses. Without the teacher-mixed sampling strategy in Section \ref{sec:rl_optim}, the distilled model outputs short responses (Case \#1) or simply repeats the input (Cases \#2).}
    \label{tab:cases}
\end{table}

\end{document}